\documentclass[conference]{IEEEtran}
\IEEEoverridecommandlockouts
\usepackage{cite}
\usepackage{amsmath,amssymb,amsfonts}
\usepackage{algorithmic}
\usepackage{graphicx}
\usepackage{subcaption}
\usepackage{cleveref}

\AtBeginDocument{%
  }

\begin{document}

\title{Multi-stage Factorized Spatio-Temporal Representation for RGB-D Action and Gesture Recognition}

\author{
    \IEEEauthorblockN{Yujun Ma$^{1,2*}$, Benjia Zhou$^{3*}$, Ruili Wang $^{1,2\dag}$, Pichao Wang$^4$}
    \IEEEauthorblockA{$^1$ Dalian Maritime University, Dalian, China}
    \IEEEauthorblockA{$^2$ Massey University, Auckland,
    New Zealand}
    \IEEEauthorblockA{$^3$ Macau University of Science and Technology, Macau, China}
    \IEEEauthorblockA{$^4$ Amazon Prime Video, Seattle, Washington, U.S.A}
    \thanks{*Both authors contributed equally to this research.}
    \thanks{$\dag$ Corresponding author.}
}

\maketitle
\begin{abstract}
 RGB-D action and gesture recognition remain an interesting topic in human-centered scene understanding, primarily due to the multiple granularities and large variation in human motion. Although many RGB-D based action and gesture recognition approaches have demonstrated remarkable results by utilizing highly integrated spatio-temporal representations across multiple modalities (\emph{i.e.,} RGB and depth data), they still encounter several challenges. Firstly, vanilla 3D convolution makes it hard to capture fine-grained motion differences between local clips under different modalities. Secondly, the intricate nature of highly integrated spatio-temporal modeling can lead to optimization difficulties. Thirdly, duplicate and unnecessary information can add complexity and complicate entangled spatio-temporal modeling. To address the above issues, we propose an innovative heuristic architecture called Multi-stage Factorized Spatio-Temporal (MFST) for RGB-D action and gesture recognition. The proposed MFST model comprises a 3D Central Difference Convolution Stem (CDC-Stem) module and multiple factorized spatio-temporal stages. The CDC-Stem enriches fine-grained temporal perception, and the multiple hierarchical spatio-temporal stages construct dimension-independent higher-order semantic primitives. Specifically, the CDC-Stem module captures bottom-level spatio-temporal features and passes them successively to the following spatio-temporal factored stages to capture the hierarchical spatial and temporal features through the Multi-Scale Convolution and Transformer (MSC-Trans) hybrid block and Weight-shared Multi-Scale Transformer (WMS-Trans) block. The seamless integration of these innovative designs results in a robust spatio-temporal representation that outperforms state-of-the-art approaches on RGB-D action and gesture recognition datasets.

\end{abstract}




\maketitle

\section{Introduction}

Action and gesture recognition has garnered significant attention in video understanding owing to its wide-ranging application scenarios, such as intelligent driving, human-machine interaction, and virtual reality \cite{wang2018rgb} \cite{liu2022spatial} \cite{miao2017multimodal}. This field has made significant progress by leveraging a wide range of data representations, such as visual appearance, skeleton, depth, and optical flows \cite{gao2022focal} \cite{zhou2021adaptive} \cite{wang2021multi}. Among those, the RGB-D-based action and gesture recognition has drawn lots of interest due to its strong adaptability to dynamic circumstances and complex backgrounds \cite{wang2017large}. 
Compared to the RGB modality, the depth modality is less sensitive to illumination, invariant to color and texture changes \cite{li2017large} and can provide detailed 3D structural information about the scene \cite{zhou2021regional}.

\begin{figure*}[htbp]
\centering
\includegraphics[height=4.3cm]{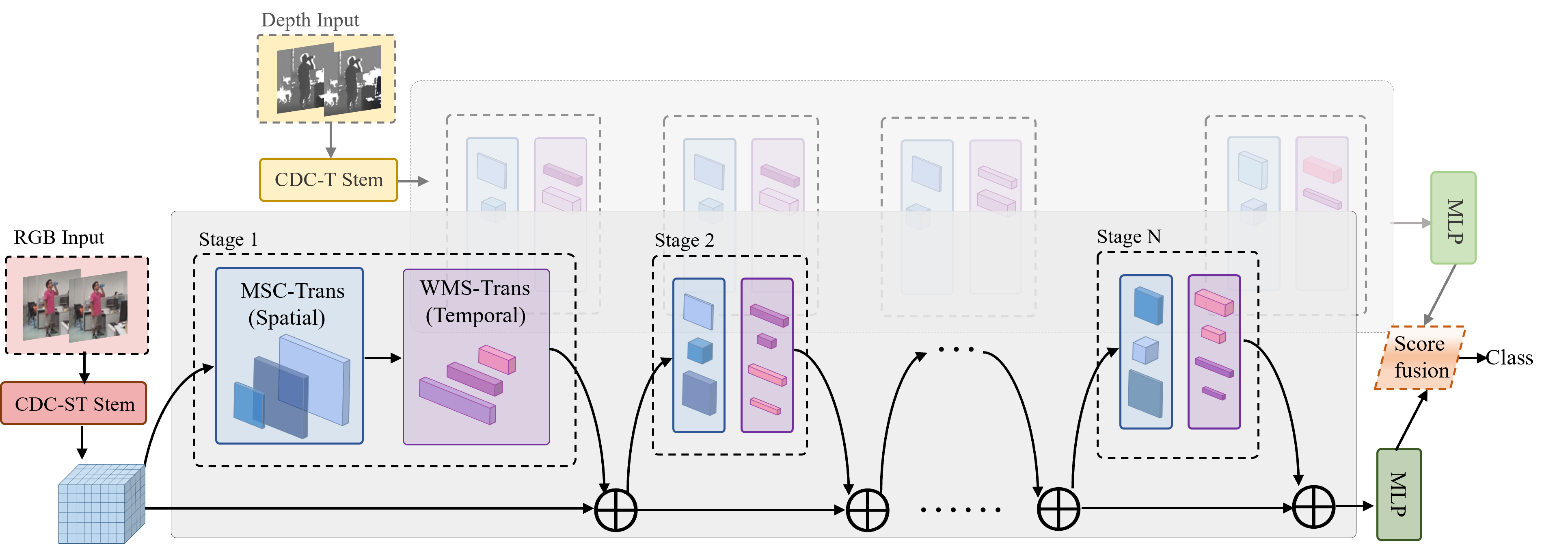}
\caption{Illustration of the proposed MFST model: CDC-Stem, multiple spatio-temporal stages. Each stage starts with MSC-Trans block for spatial features, followed by WMS-Trans layers for temporal features. Residual connections optimize the model, and $ \oplus $ indicates element-wise addition.}

\label{Fig1}
\end{figure*}

The existing RGB-D action and gesture recognition methods \cite{cheng2021cross} \cite{liu2022dual} \cite{zhang2017learning} commonly employ convolutional neural networks (CNNs) and recurrent neural networks (RNNs) to extract features from RGB and depth cues \cite{narayana2018gesture}.  However, in recent years, Transformer-based methods \cite{li2021trear} \cite{ahn2023star} \cite{liu2023cross} have achieved surprising performance for modeling long-range dependencies regarding temporal dynamics for RGB-D action and gesture recognition. Transformer \cite{vaswani2017attention} is suitable for processing sequential data as it can effectively model the relationships among different frames, which enables the model to recognize and interpret the same motion occurring at different time indexes.  However, most Transformer-based methods \cite{arnab2021vivit} \cite{plizzari2021skeleton} \cite{yang2022recurring} process spatial and temporal information using the same strategy or combine spatio-temporal information without explicitly analyzing the differences between the spatial and temporal dimensions in RGB-D data.

Specifically, there are still some problems in the following three aspects. (i) \textbf{Fine-grained motion differences} might be overlooked when employing vanilla 3D convolution (\emph{e.g.}, C3D\cite{li2016large} and I3D \cite{carreira2017quo} ), which operates on the entire input volume simultaneously, without considering the subtle motion differences that may arise within local clips or across different modalities. This can limit the ability of the model to distinguish between similar actions or events that differ only in subtle ways. (ii) \textbf{Optimization difficulties} often arise when working with limited video data, primarily due to the highly interdependent and intertwined nature of spatio-temporal modeling. The intricate relationships between the spatial and temporal information can make optimizing the model parameters challenging, resulting in reduced performance. (iii) \textbf{Duplicate and unnecessary information} can create additional complexity and make it difficult to deal with in entangled spatio-temporal modeling. To solve the above mentioned problems, some decoupled models \cite{neimark2021video} \cite{zhang2021vidtr} \cite{bulat2021space} have been proposed, which model spatial and temporal representations separately. For example, Bertasius \emph{et al}.\cite{bertasius2021space} presented a divided space-time attention model for action recognition, which separately applied temporal and spatial attention within each block. However, such divided space-time attention could potentially result in losing essential spatio-temporal relationships since there is no explicit mechanism for spatial and temporal feature interaction or spatio-temporal feature coupling. Later, Zhou \emph{et al}.\cite{zhou2022decoupling} proposed to utilize a distillation based recouple strategy to establish the space-time dependency of decoupled spatial and temporal representations for RGB-D based action and gesture recognition. However, such decoupling and recoupling structures could reintroduce bias or errors and require additional computation. Furthermore, distilling the inter-frame correlations from the time dimension into the space dimension could potentially guide the model prone to temporal information. (iiii) \textbf{Insufficient exploration} of RGB-D multi-modality for action and gesture recognition based on videos. Most existing approaches focus on RGB action and gesture recognition, which may not be sufficient to capture a person's subtle movements and poses during an action. However, cost-effective depth sensors like Microsoft Kinect can capture depth information in addition to color information, enabling more accurate tracking of body movements and poses, providing additional pixel-level positional cues of the scene for recognizing actions and gestures. Thus, there is still a significant research gap in exploring effective multi-modal spatio-temporal representation for dynamics RGB-D action and gesture recognition.

Given the above-mentioned concerns, as illustrated in Figure \ref{Fig1}, we introduce a new Multi-stage Factorized Spatio-Temporal (MFST) learning architecture for RGB-D-based action and gesture recognition. Broadly, it begins with a lightweight 3DCDC-based \cite{yu2021searching} convolutional Stem (CDC-Steam) module to capture fine-grained spatio-temporal features. Then, inspired by the computationally efficient manner of CNN's hierarchical structure, we partition the spatio-temporal learning process into multiple factorized spatio-temporal stages with residual connections to learn hierarchical spatial and temporal features via the proposed Multi-Scale Convolution and Transformer (MSC-Trans) hybrid block and Weight-shared Multi-Scale Transformer (WMS-Trans) block.


In more detail, as shown in Figure \ref{Fig1}, the gist of our proposed model is composed of three aspects: (1) \textbf{Convolutional lightweight Stem}. Firstly, we introduce a Central Difference Convolutional (CDC) \cite{yu2021searching} based Stem to capture bottom-level features (\emph{e.g.}, patterns and colors) that form fundamental structures in video frames (\emph{e.g.}, corners and lines). More specifically, we use the Spatio-Temporal CDC based Stem (CDC-ST Stem) to process the RGB data and the Temporal CDC based Stem (CDC-T Stem) to process the depth data. This consideration of stems from the fact that the former is sensitive to both temporal and spatial priors, while the latter is only sensitive to temporal priors. (2) \textbf{Multi-stage hierarchical structure}. The multi-stage process contains spatial feature embeddings and temporal embeddings simultaneously. This hierarchical manner allows the model to model local spatio-temporal contexts from low-level edges and build up to higher-order semantic primitives using a multi-stage hierarchy structure, similar to Convolutional Neural Networks (CNNs). Concurrently, temporal and spatial dependencies have also been implicitly modeled in a single stage. (3) \textbf{Factrized spatio-temporal learning}. As mentioned before, each stage begins with a Multi-Scale Convolution and Transformer (MSC-Trans) hybrid spatial block, and a Weight-shared Multi-Scale Transformer (WMS-Trans) temporal block comprises the remainder of each stage. Concretely, the MSC-Trans spatial block consists of a stack of inception-based multi-scale spatial features learning layers to extract hierarchical local spatial features, coupled with a Transformer that captures global spatial features. The WMS-Trans temporal block is composed of multiple Transformer layers for hierarchical local fine-grained and global coarse-grained temporal feature learning. Specially, we adopt a Transformer structure based on Kvt\cite{wang2022kvt}, which reduces the redundant marginal information in extracted temporal features.

Our work is inspired by the De-Recouple model [50], but we introduce the MFST approach, which differs in spatial-temporal representation. Our method employs a fine-grained multi-stage process, capturing spatio-temporal features individually in each stage. In contrast, De-Recouple uses a global way, potentially leading to temporal dominance due to asymmetries in the modeling capabilities of sub-networks. Additionally, we propose a weight-shared single Transformer branch for more efficient temporal multi-scale feature modeling compared to their multi-branch approach. With these designs, our MFST model effectively learns spatio-temporal features within each modality for action and gesture recognition. To the best of our knowledge, we are the first to propose an efficient factorized spatio-temporal representation at the fine-grained stage level for RGB-D action and gesture recognition rather than a coarse-grained model level.


\begin{figure*}[htbp]
\centering
\includegraphics[height=5.5cm]{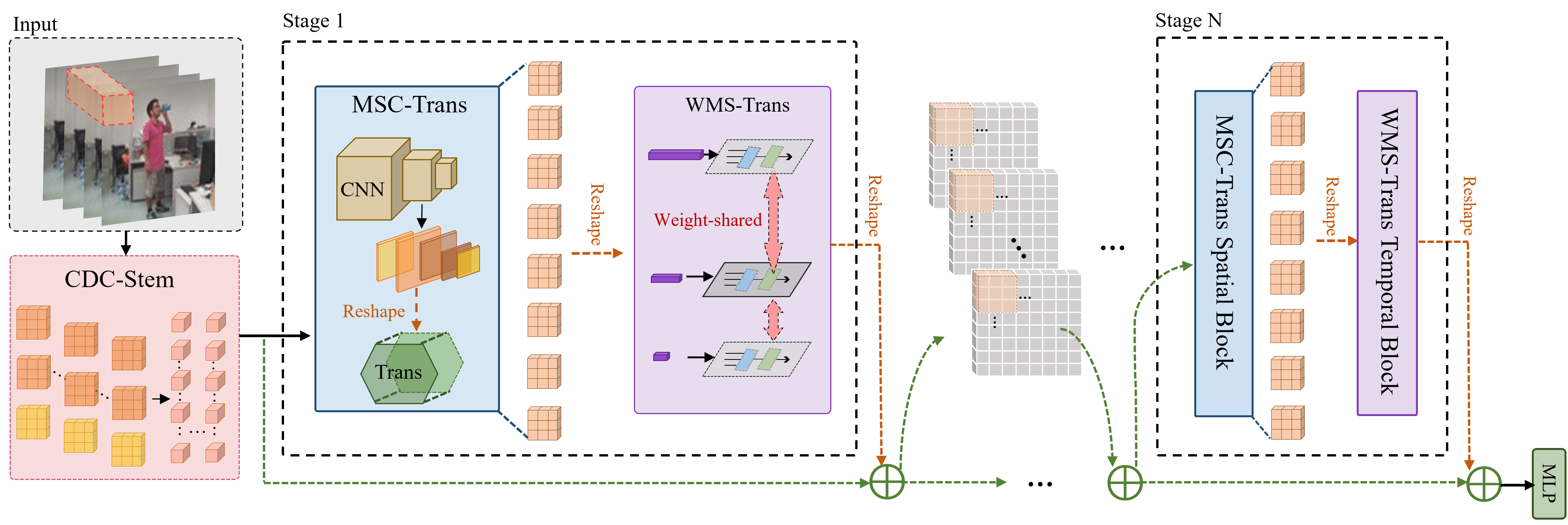}
\caption{Illustration of our proposed MFST model for RGB-D action and gesture recognition includes CDC-Stem, multiple stages with MSC-Trans hybrid spatial blocks, and WMS-Trans temporal blocks. Residual connections are utilized between consecutive stages to optimize the model, and $ \oplus $ indicates element-wise addition.}

\label{Fig2}
\end{figure*}


\vspace{-0.1cm}
\section{Related work}

\subsection{RGB-D Data based Action and Gesture Recognition}

With the recent advancements in cost-effective RGB-D sensors, such as Microsoft Kinect and Asus Xtion, there has been a growing interest in RGB-D-based action and gesture recognition. This is mainly due to the fact that the additional dimension (depth) is unaffected by changes in illumination and provides comprehensive 3D structural information of the motion \cite{zhang2018attention}. Recent works in RGB-D gesture and action recognition have proposed different techniques to capture and enhance temporal and spatial information from RGB-D data. Wang \emph{et al}. \cite{wang2017scene} utilized  scene flow for compact RGB-D representation learning, which considered the modalities as one entity. Later, Yu \emph{et al}. \cite{yu2021searching} proposed a temporal enhanced multi-modality model for gesture recognition, which utilized three different types of 3D central difference convolution to capture rich temporal dynamics information. Wang \emph{et al}. \cite{wang2018cooperative} proposed a single convolutional neural network based on cooperative training for RGB-D action recognition, which enhanced the discriminative power of the deeply learned features. Cheng \emph{et al}.\cite{cheng2021cross} proposed a cross-modality compensation model for RGB-D action recognition, which employed a compensation block to capture interaction information between different modalities. Liu \emph{et al}.\cite{liu2022dual} presented a dual-stream Transformer model for RGB-D action recognition that incorporates dual streams and a cross-modality fusion layer to extract spatio-temporal features simultaneously. Unlike global-based spatio-temporal decoupling methods \cite{zhang2021vidtr} \cite{zhou2022decoupling}, we focus on decomposing spatio-temporal dependencies in each local stage. Our approach can achieve the goal of spatio-temporal decoupling and reduce the risk of spatio-temporal connection weakening caused by the complete decoupling of space and time. 

\subsection{Disentangled Spatio-temporal Representation}

Disentangled spatio-temporal models decouple the spatial and temporal information in videos, enabling the use of spatial and temporal architectures optimized for their respective domains. Shi \emph{et al}. \cite{shi2020decoupled} present a decoupled spatio-temporal attention network for skeleton-based action recognition, emphasizing the specific characteristics of space/time and different motion scales. Bertasius \emph{et al}. \cite{bertasius2021space} introduced a divided space and time attention Transformer model for RGB action recognition, which first computed self-attention across the spatial dimensions, followed by computing self-attention across the temporal dimension. Zhang \emph{et al}.\cite{zhang2021vidtr} proposed performing spatial and temporal attention for RGB action recognition, respectively, by decoupling the 3D self-attention to a spatial attention and a temporal attention.  Liu \emph{et al}. \cite{liu2021decoupled} proposed a disentangled spatio-temporal Transformer for video inpainting. The model attended to temporal features on different frames at the same spatial pixels and then to similar background textures on the same frame at all spatial positions. However, coarse-grained or global-level model decoupled spatio-temporal networks have a limitation in effectively incorporating dependencies between spatial and temporal information, which can result in incomplete decoupling. In this study, we aim to acquire both spatial and temporal characteristics, as well as their dependencies, for video-based RGB-D action and gesture recognition. To achieve this, we propose a hierarchically multi-stage factorized spatio-temporal representation that incorporates residual connections.

\section{Proposed Method}

\subsection{Overview}

In this paper, we propose a hierarchical multi-stage factorized spatio-temporal feature representation learning architecture for RGB-D action and gesture recognition. Multi-stage hierarchy design borrowed from CNNs is adopted, where four stages in total are used in this work (described in Section 3.2). The proposed MFST model comprises a 3D Central Difference Convolutional Stem (CDC Stem) module (detailed in Section 3.3), a sequence of factorized spatio-temporal feature learning stages (described in Section 3.4), and an MLP-based prediction head. 



\subsection{Hierarchical Multi-stage Factorized Spatio-Temporal Representation }

The MFST model (Figure \ref{Fig2}) comprises a 3D Central Difference Convolutional Stem (CDC-Stem) module as its first step. This module generates visual embeddings that are fed into cascaded factorized spatio-temporal stages, consisting of four stages in total. To capture multi-scale spatial features both locally and globally, each stage uses a hybrid spatial block (MSC-Trans) that consists of a Multi-Scale Convolution module followed by a Transformer. Then, the spatial features obtained from each stage are input into a Weight-shared Multi-Scale Transformer block (WMS-Trans) to capture multi-scale temporal features. Additionally, we also introduce the residual connection \cite{he2016deep} between consecutive stages to ensure stable gradients. Finally, an MLP layer makes the final predictions based on the output obtained from the final stage.

 Let $V\ \in \mathbb{R}^{T\times 3\times H\times W}$, where \emph{T} is the length of the video and \emph{H} and \emph{W} represent the height and width of each frame, respectively. The embeddings \emph{Z} after the first stage can be formally expressed as:

\begin{equation}
Z = WSM\text{-}Trans(MSC\text{-}Trans(CDC\text{-}Stem(V))).
\end{equation}

As depicted in Figure \ref{Fig2}, the output of the MSC-Trans block is reshaped to serve as input for the WMS-Trans block.

The factorized spatio-temporal representation stage combines the MSC-Trans spatial block and WMS-Trans temporal block, enabling the simultaneous learning of visual patterns that incorporate both local spatial information and higher-order temporal information. Moreover, integrating these two types of blocks within a single stage allows the model to effectively capture spatio-temporal context at different levels of complexity. The factorized spatio-temporal representation stage also facilitates the effective modeling of spatio-temporal dependencies without recoupling or reconstructing spatio-temporal representations, unlike the approach\cite{zhou2022decoupling}.

\subsection{\textbf{Central Difference Convolutional Stem}}

We propose a lightweight Central Difference Convolutional Stem (CDC-Stem) module that embeds bottom-level visual features such as lines, edges, and corners before passing them to subsequent multi-stage representations. The Stem is constructed using the first five layers of the Inflated 3D Convolutional Neural Network (I3D) \cite{carreira2017quo}, similar to the design used in \cite{szegedy2017inception}. However, vanilla 3D convolution makes it hard to capture fine-grained motion differences between local clips across different modalities. To further enhance the spatio-temporal representation and fully exploit rich local motion information, we adopt the 3D Central Difference Convolution (3D-CDC) \cite{yu2021searching} in our Stem module.

Specifically, we replace the vanilla 3D operation with 3D-CDC in the Stem and employ different CDC operations for RGB modality and Depth data. As illustrated in Figure \ref{Fig3}, the 3D-CDC module consists of two distinct steps: sampling and aggregation. The sampling step is similar to vanilla 3D convolution, while the aggregation step differs. Figure \ref{Fig3}(a) shows the Spatio-Temporal Central Difference Convolution (CDC-ST) in the Stem module for the RGB modality, and Figure \ref{Fig3}(b) shows the Temporal Central Difference Convolution (CDC-T) in the Stem module for the depth modality.

The input tensor \emph{V} is first passed through a CDC-Stem module. In general, the CDC-Stem module is written as follows:

\begin{equation}
E\ =\ CDC\text{-}Stem( V, \Psi) ,\ E\ \in \mathbb{R}^{T\times C\times H\times W},
\end{equation}
where $\Psi$ indicates the learnable parameter matrices. After passing the input tensor $\emph{V}$ through the CDC-Stem module, the resulting embedded spatio-temporal features $\emph{E}$ are further processed by hierarchical factorized spatio-temporal feature learning stages.




\subsubsection{\textbf{CDC-ST Stem for RGB}}

We adopt CDC-ST in the Stem module for the RGB modality because RGB data provides detailed appearances, such as local shape changes of the hands and fingers, which are better suited to spatial gradient cues and appearance details. CDC-ST combines spatio-temporal gradient information into a single 3D convolution operator, effectively capturing subtle changes across the frame sequence. Thus, our proposed CDC-ST Stem leverages the rich local motion information to enhance the spatial-channel representation, which is critical for identifying the fine-grained temporal dynamics of motions and differentiating between similar actions and gestures.


Explicitly, the vanilla 3D convolution operation involves sampling a local receptive field cube $\mathcal{Z}$ over the input feature map \emph{x} and aggregating the sampled values via a weighted summation with learnable weights \emph{w}. The output feature map ${y_{3D}}$ can be formulated as:

\begin{equation}
y_{3D}( p_{0}) =\sum _{p_{n} \in \mathcal{Z}} w( p_{n}) \cdot x( p_{0} +p_{n}),
\end{equation}
where $p_{0}$ indicates the current location on both input and output feature maps while $p_{n}$ enumerates the locations in $\mathcal{Z}$. The output feature map of CDC-ST can be formulated as:

\begin{equation}
y_{CDC-ST}( p_{0}) =\ \sum _{p_{n} \in \mathcal{Z}} w( p_{n}) \cdot ( x( p_{0} +p_{n}) -x( p_{0})).
\end{equation}

To enhance the robustness and discriminative modeling capacity in the proposed CDC-ST Stem module, we utilized the setting described in \cite{yu2021searching}, which involves a combination of vanilla 3D convolution and 3D-CDC-ST. The generalized 3D-CDC-ST operation can be formulated as follows:

\begin{equation}
y_{3D-CDC-ST}( p_{0}) =\theta \cdot y_{CDC-ST}( p_{0}) +( 1-\theta ) y_{3D}( p_{0}),
\end{equation}
where the hyperparameter $\theta$ ranges from $\emph{0}$ to $\emph{1}$, determines the trade-off between the contribution of intensity-level and gradient-level information.

\subsubsection{\textbf{CDC-T Stem for Depth}}

We adopt CDC-T in the Stem module for the depth modality because depth data directly captures the distance between objects and the camera, which is more sensitive to temporal changes in the environment than RGB data. However, depth data is often noisier and less detailed than RGB data, making it challenging to extract accurate temporal information. By proposing CDC-T Stem, we can effectively capture central temporal differences and filter out some of the noise in the depth data, resulting in a sequence of filtered depth frames that are more sensitive to dynamic temporal changes.

\begin{figure}[htbp]
\centering
\includegraphics[height=8cm]{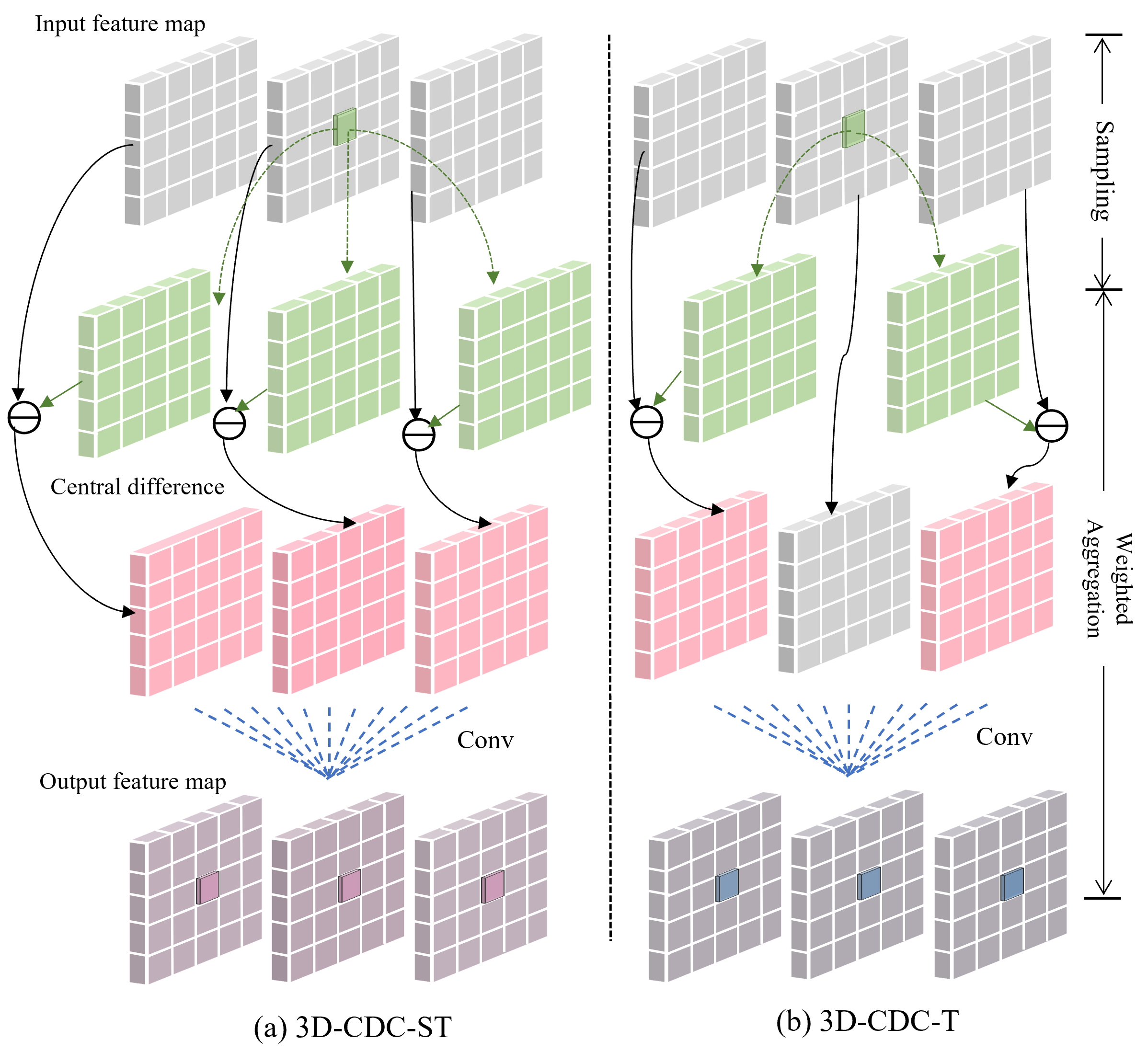}
\caption{(a) 3D-CDC-ST incorporates central difference information from the entire local spatio-temporal regions. (b)3D-CDC-T only utilizes central difference clues from the local spatio-temporal regions of the adjacent frames.}
\label{Fig3}
\end{figure}

As shown in Figure \ref{Fig3}(b), in 3D-CDC-T operation, the sampled local receptive field cube $\mathcal{Z}$ is divided into two regions: 1) the region in the current time step $\mathcal{T}'$, and 2) the regions in the adjacent time steps $\mathcal{T}''$. In the context of a 3D-CDC-T, the central difference term is only computed from the region $\mathcal{T}''$. Thus the generalized 3D-CDC-T can be formulated as follows:

\begin{equation}
y_{3D-CDC-T}( p_{0}) =y_{3D}( p_{0}) +\theta \cdot \left( -x( p_{0}) \cdot\sum _{p_{n} \in \mathcal{T} ''} w( p_{n})\right).
\end{equation}

The results of the ablation experiments in Section 4.3.2 confirm that 3D-CDC-ST is more effective in processing RGB data, while 3D-CDC-T performs better in handling depth data. The experimental results align with our hypothesis that 3D-CDC-T is better suited for depth data due to its ability to reduce sensor noise and pre-processing artifacts between frames of depth data. Moreover, processing depth data requires more emphasis on temporal reasoning context rather than spatial gradient cues and appearance details, which is why 3D-CDC-ST is better suited for RGB data.

\subsection{Factorized Spatio-Temporal Representation}

Our proposed model involves the factorization of spatio-temporal representation in each stage. Each stage commences with a Multi-Scale Convolution and Transformer (MSC-Trans) hybrid spatial block, followed by a Weight-shared Multi-Scale Transformer (WMS-Trans) temporal block that comprises the remainder of the stage.

\subsubsection{\textbf{MSC-Trans block}}

\begin{figure}[htbp]
\centering
\includegraphics[height=2.2cm]{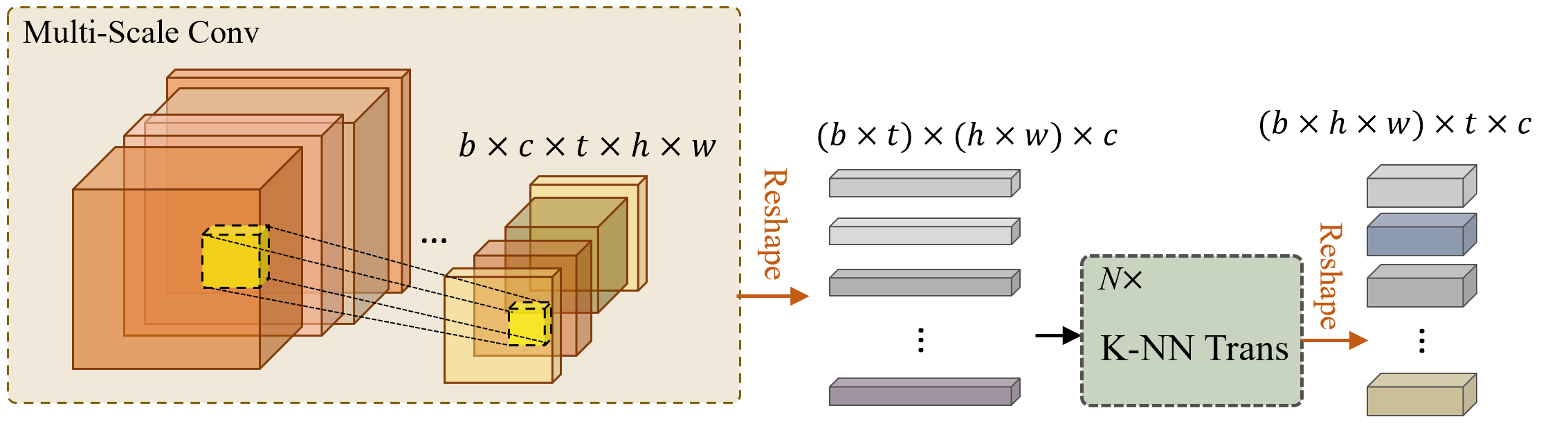}
\caption{The MSC-Trans block is composed of two main parts: the Multi-Scale Convolution (MSC) layers and a K-NN Transformer.}
\label{Fig4}
\end{figure}

As shown in Figure \ref{Fig4}, the MSC-Trans spatial block consists of a stack of inception-based multi-scale spatial features learning layers to extract hierarchical local spatial features, coupled with a k-NN Transformer \cite{wang2022kvt} that captures global spatial features. Multi-Scale Convolution (MSC) operation comprises a space-centric 3D Inception Module and a 3D Max Pooling layer, which models local multi-scale spatial features, as denoted by: 

\begin{equation}
F_{MSC} =\ MaxPool( InC_{1\times k\times k}( X,\theta )),
\end{equation}
where $InC_{1\times k\times k}( X,\theta )$ refers to the Inception Module with a kernel size of $1\times k\times k$. The notion of capturing hierarchical spatial features arises from the recognition that spatial patterns vary at multiple scales when an action takes place.

The multi-scale local spatial feature from MSC is then reshaped into a size that can be accepted by the Transformer. To reduce the redundant marginal information in captured temporal features, here we utilize a Transformer structure based on K-NN multi-head self-attention layer \cite{wang2022kvt} for global spatial feature modeling. Therefore, the modeling process of the K-NN Transformer can be computed through:

\begin{equation}
L_{Trans} =MLP( LN( MSA_{knn}( F_{MSC}))),
\end{equation}
where $F_{MSC}$ denotes the output of the MSC operation, $ MSA_{knn}$ refers to k-NN multi-head self-attention layer and \emph{LN} represents layer normalization. In order to maintain a balance between the MSC-Trans block and WMS-Trans block and prevent the factored spatio-temporal stage from being biased towards any dimension, we employ four Transformer layers with a head number of eight in the proposed MSC-Trans block.

\subsubsection{\textbf{WMS-Trans block}}

Video data contains both spatial and temporal information, and existing approaches typically use 2D convolutions to capture spatial information and 3D components in the higher layers to extract temporal information \cite{yu2019weakly} \cite{li2020spatio}. Building on this concept, we propose a Weight-shared Multi-scale Transformer (WMS-Trans) temporal block after the MSC-Trans spatial block. The WMS-Trans block comprises a Transformer branch that performs hierarchical local fine-grained and global coarse-grained temporal feature learning.

\begin{figure}[htbp]
\centering
\includegraphics[height=4.8cm]{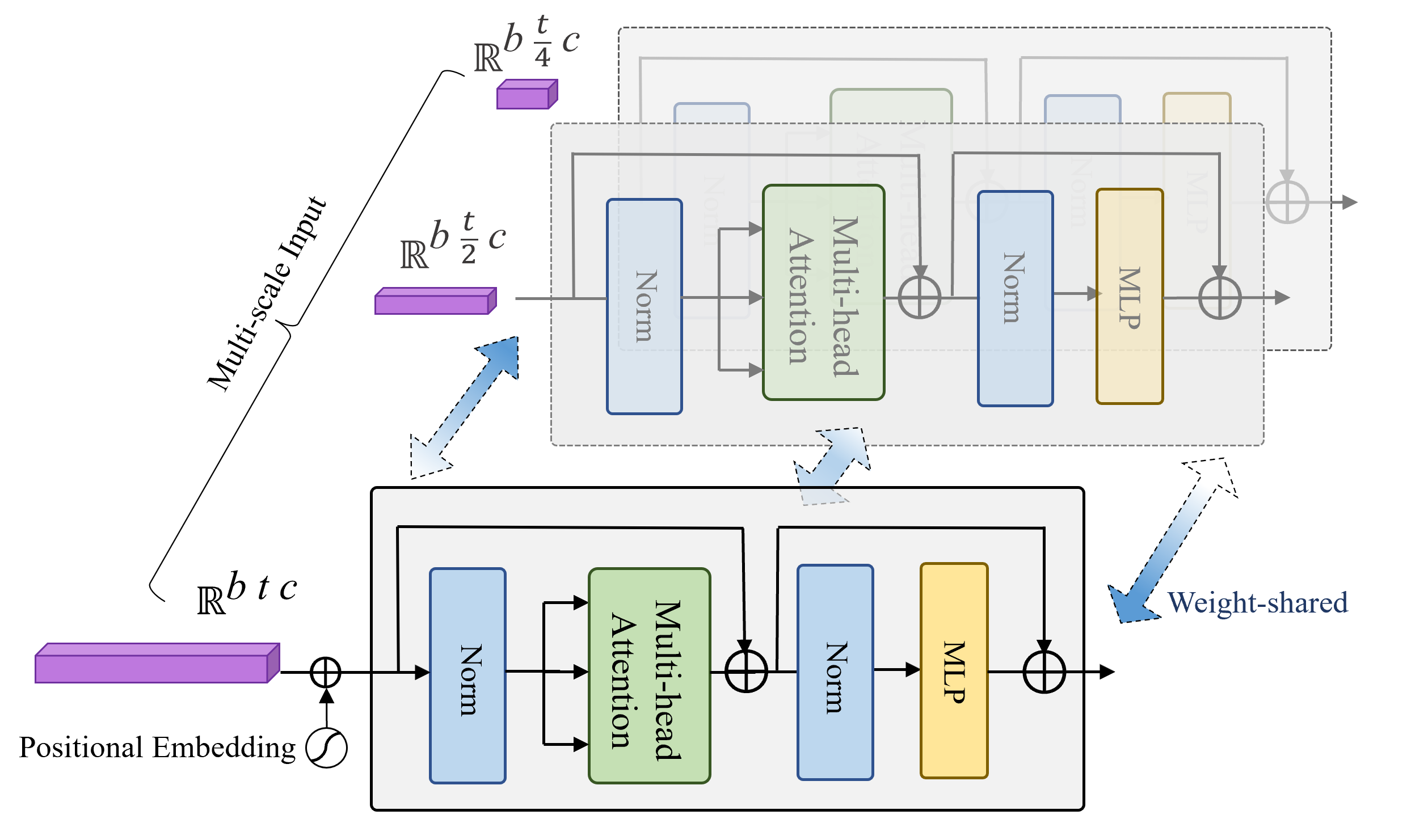}
\caption{Illustration of the WMS-Trans block, consisting of a Transformer branch with multi-scale temporal dimension inputs, and a weight-shared strategy is applied.}
\label{Fig5}
\end{figure}

To obtain multiple temporal scales, we reshape the multi-scale spatial features from the previous MSC-Trans block into various temporal dimensions. As shown in Figure \ref{Fig5}, the Transformer branch in the WMS-Trans block models multi-scale temporal features by handling different temporal dimensions of the input sequence. For example, the multi-scale input can be represented as ($\mathbb{R}^{b,t,c}$), ($\mathbb{R}^{b,\frac{1}{2}t,c}$), and ($\mathbb{R}^{b,\frac{1}{4}t,c}$), which cover short-term to long-term temporal dependencies. A weight-shared strategy enables saving the number of parameters and promotes interaction between different temporal scales. The WMS-Trans block can effectively learn a more generalizable representation of the input embeddings by sharing the same parameters across different temporal scales. Our weight-shared multi-scale Transformer differs from the De+Recouple model \cite{zhou2022decoupling}. They use a multi-branch Transformer for capturing multi-scale temporal information without weight-shared, which is computationally heavier and may neglect dependencies among different scales of temporal information.

Furthermore, we have incorporated positional embeddings into the input embeddings of the WMS-Trans block. As shown in Figure \ref{Fig5}, the positional encodings have the same dimensions as the input embeddings, allowing them to be added together. We have utilized sine and cosine functions with varying frequencies to achieve this:

\begin{equation}
PE( pos,2i) =sin\left( pos/10000^{2i/d_{model}}\right),
\end{equation}

\begin{equation}
PE( pos,2i+1) =cos\left( pos/10000^{2i/d_{model}}\right),
\end{equation}
where \emph{pos} is the position and \emph{i} is the dimension. The WMS-Trans block can learn spatial relationships and capture meaningful patterns in the input embeddings by adding positional embedding. To sum up, the WMS-Trans block is capable of capturing temporal features at different time scales hierarchically, allowing it to model complex and wide range temporal patterns in the input embeddings.

\section{Experiments}

\subsection{Implementation Details}

All experiments were conducted using PyTorch and were run on 4 × A100 GPUs. Input sequences were randomly or center-cropped to $224\times224$ during both training and inference. We utilized SGD as the optimizer with a weight decay of 0.0003 and a momentum of 0.9. The learning rate was linearly ramped up to 0.01 during the first three epochs and then decayed with a cosine schedule. The training process lasted 100 epochs, and we adopted MixUp\cite{zhang2017mixup} as a data augmentation strategy.

We evaluated our MFST baseline model (MFST-BM) on the NTU RGB-D dataset, which does not incorporate CDC, residual connections, and weight-shared Transformers. However, it still employs a hierarchical multi-stage factorized spatio-temporal approach using vanilla 3D convolutions and Transformers. Furthermore, we conducted experiments with two variations of our model that incorporate CDC, residual connections, and weight-shared Transformers: MFST-Base, which has three stages, and MFST-Large, which has four stages. In the basic configuration, we used four Transformer layers ($L=4$) in the temporal feature learning block and eight heads ($N_{H} = 8$). Unless specified otherwise, we refer to this setting as the basic configuration of our network.

\subsection{Comparison with State-of-the-art Methods}

The proposed method achieves state-of-the-art (SOTA) performance on both large-scale RGB-D action and gesture recognition datasets (\emph{e.g.}, NTU RGB-D, and IsoGD). More SOTA comparison performance can be found in the supplementary material.

\subsubsection{\textbf{NTU RGB-D Dataset}}

NTU RGB-D \cite{shahroudy2016ntu} is a highly utilized large-scale indoor action recognition dataset that includes over 56 thousand action clips and 4 million frames across 60 action classes. The dataset consists of clips performed by 40 volunteers captured in a constrained laboratory setting using three KinectV2 cameras, each providing a different viewpoint. NTU RGB-D presents a significant challenge due to the considerable intra-class and viewpoint variations. Many recent studies have utilized the skeleton information to perform 2D/3D convolutions for action recognition. However, skeleton data provide limited information about the human body and its movements, are more sensitive to occlusions, and inability to capture fine-grained details. In this work, we focus solely on the RGB and depth modalities for action recognition. Our results, as presented in Table 1, demonstrate SOTA performance on both the Cross-View (CV) and Cross-Subject (CS) protocols.  Compared to the SOTA skeleton-based method Ta-CNN\cite{xu2022topology}, our proposed MFST-Large model achieves approximately 4.2\% on the CS protocol and 3.4\% on the CV protocol. Compared to the SOTA RGB-D based method De+Recouple\cite{zhou2022decoupling}, our proposed MFST-Large model achieves approximately 0.4\% on the CS protocol and 0.9\% on the CV protocol. Our results demonstrate superior performance over the De+Recouple\cite{zhou2022decoupling} model on both the RGB and depth modalities. The proposed MFST model shows strong motion perception abilities and robustness to noisy backgrounds, as evidenced by the comparable or superior performance of the depth modality.

\begin{table}[]
\caption{Comparison of the SOTA methods on NTU RGB-D.}
\resizebox{1\linewidth}{!}{
\begin{tabular}{lllll}
\hline
Methods                                & Publisher & Modality       & CS (\%)    & CV(\%)     \\ \hline
S-GCN\cite{cheng2020skeleton}          & CVPR20    & SKL            & 90.7       & 69.2       \\
DC-GCN\cite{cheng2020decoupling}       & ECCV20    & SKL            & 90.8       & 69.4       \\
MST-GCN\cite{chen2021multi}            & AAAI21    & SKL            & 91.5       & 96.6       \\
ResGCN\cite{song2020stronger}          & ACMMM20   & SKL            & 90.9       & 96.0       \\
STST\cite{zhang2021stst}               & ACMMM21   & SKL            & 91.9       & 96.8       \\
Ta-CNN\cite{xu2022topology}            & AAAI22    & SKL            & 90.4       & 94.8       \\ \hline
STAR-Trans\cite{ahn2023star}           & CVPR23    & SKL+RGB        & 92.0       & 96.5       \\
D-Bilinear\cite{hu2018deep}            & ECCV18    & RGB-D+SKL      & 85.4       & 90.7       \\
Coop-CNN\cite{wang2018cooperative}     & AAAI18    & RGB+DEP        & 86.4       & 89.1       \\
P4Trans\cite{fan2021point}             & CVPR21    & Points         & 90.2       & 96.4       \\
De+Recouple\cite{zhou2022decoupling}   & CVPR22    & RGB            & 90.3       & 95.4       \\
De+Recouple\cite{zhou2022decoupling}   & CVPR22    & DEP            & 92.7       & 96.2       \\  
De+Recouple\cite{zhou2022decoupling}   & CVPR22    & RGB+DEP        & 94.2       & 97.3       \\  \hline
Ours(MFST-Large)                      & -         & RGB            & \textbf{92.1}       & \textbf{95.8} \\
Ours(MFST-Large)                     & -         & Depth          & \textbf{93.8}       &  \textbf{97.3} \\
Ours(MFST-Large)                           & -         & RGB+DEP        &  \textbf{94.6}     &  \textbf{98.2}   \\ \hline
\end{tabular}
}
\end{table}



\subsubsection{\textbf{Chalearn IsoGD Dataset}}

The Chalearn IsoGD dataset \cite{wan2016chalearn} comprises 47,933 RGB-D gesture videos categorized into 249 different types of gestures, and 21 individuals performed these videos. This dataset is considered more challenging due to two main factors: firstly, it covers gestures in various fields and ranges of motion, from delicate fingertip movements to broad arm swings. Secondly, many of the gestures exhibit high similarity. However, as shown in Table 2, our model has demonstrated impressive performance on this dataset despite these challenges. In terms of RGB-D gesture recognition, our MFST-Large model achieves more than 2\% accuracy improvement compared to the SOTA methods using RGB and depth modalities. This success can be attributed to utilizing hierarchical and consolidated features learned through multi-stage factorized spatio-temporal representation coupled with multi-scale feature modeling, which can effectively capture subtle dynamic differences between similar gestures.

\begin{table}[]
\centering
\caption{Comparison of the SOTA methods on IsoGD.}
\begin{tabular}{lll}
\hline
Methods       & Modality  & Accuracy(\%) \\ \hline
3DDSN\cite{duan2018unified}                  & RGB       & 46.08        \\
AttebtionLSTM\cite{zhu2019redundancy}        & RGB       & 57.42        \\
NAS\cite{yu2021searching}                    & RGB       & 58.88        \\
De+Recouple\cite{zhou2022decoupling}         & RGB       & 60.87        \\
Ours(MFST-Large)                             & RGB       & \textbf{61.26}      \\ \hline
AttentionLSTM\cite{zhu2019redundancy}        & DEP       & 54.18        \\
3DDSN\cite{duan2018unified}                  & DEP       & 54.95        \\
NAS\cite{yu2021searching}                    & DEP       & 55.68        \\
De+Recouple\cite{zhou2022decoupling}         & DEP       & 60.17        \\
Ours(MFST-Large)                             & DEP       & \textbf{61.29}      \\ \hline
AttentionLSTM\cite{zhu2019redundancy}        & RGB+DEP   & 61.05            \\
NAS\cite{yu2021searching}                    & RGB+DEP   & 62.47        \\
De+Recouple\cite{zhou2022decoupling}         & RGB+DEP       & 66.79        \\
Ours(MFST-Large)                             & RGB+DEP    & \textbf{68.47}   \\ \hline
\end{tabular}
\end{table}




\subsection{Ablation Study and Analysis}

In the ablation study, we utilized the NTU RGB-D dataset and conducted all experiments using the Cross-Subject (CS) protocol. For additional ablation studies, we refer the readers to our supplementary material.

\subsubsection{\textbf{Effectiveness of Multi-stage Factorized Spatio-Temporal learning}}

Table 3 demonstrates that our proposed MFST-Base already surpasses De+Recouple\cite{zhou2022decoupling} model, achieving a 0.2\% increase in top-1 accuracy on the RGB data while utilizing fewer parameters and input frames. Moreover, our MFST-Large model has outperformed our MFST-Base model by 1.6\% and has also outperformed De-recouple \cite{zhou2022decoupling} by 1.8\%. This is because, unlike the other decoupled methods \cite{zhou2022decoupling}\cite{zhang2021stst}, which disentangle spatio-temporal features globally at the whole model level. Our multi-stage structure with factorized spatio-temporal learning can effectively capture local spatio-temporal contexts from low-level edges and gradually build up to higher-order semantic primitives in videos. In addition, performance continually improves as more stages use the design, validating the MFST model as an effective modeling strategy. Additional ablation experiments have been conducted on the MFST-Large model unless mentioned. 

\begin{table}[]
\centering
\caption{Ablation study on multi-stage factorized spatio-temporal learning. MFST-Base has three stages, and MFST-Large has four stages.}

\begin{tabular}{llll}
\hline
Models      & Acc(\%)    & \#params & \#frames \\ \hline
De-Recouple\cite{zhou2022decoupling} & 90.3 & 38.0MB     & 64       \\
MFST-Base   & 90.5 & 32.6MB     & 32       \\
MFST-Large  & 92.1 & 49.7MB     & 32       \\ \hline
\end{tabular}
\end{table}

\begin{table}[]
\centering
\caption{Comparison among different types of 3D convolutions in the CDC-Stem module, including vanilla 3D, 3D-CDC-ST, and 3D-CDC-T.}
\begin{tabular}{llll}
\hline
Stem-Setting & RGB(\%) & DEP(\%) &$\theta$                 \\ \hline
Vanilla-3D   & 90.4    & 91.2      & 0                     \\ \hline
3D-CDC-ST    & \textbf{92.1}    & 91.6      & 0.6                   \\
3D-CDC-T     & 89.6    & \textbf{93.8}      & 0.6                   \\ \hline
3D-CDC-ST    & 91.0    & 91.0      & 0.8                   \\
3D-CDC-T     & 89.9    & 92.4      & 0.8                   \\ \hline
\end{tabular}
\end{table}

\subsubsection{\textbf{3D Central Difference Convolution based Stem}} Table 4 presents a comparison between vanilla 3D convolution and 3D central difference convolution within the context of our proposed lightweight Stem module. We have discovered that 3D-CDC-ST performs better when applied to RGB data, while 3D-CDC-T yields better results with depth data. This is because 3D-CDC-T can help to reduce sensor noise and pre-processing artifacts in depth data by leveraging the temporal context between frames. Depth data is often noisier than RGB data, and using temporal context can help to smooth out these fluctuations \cite{yu2021searching}. On the other hand, RGB data provides information about both spatial details and temporal motion patterns. This means that spatial gradient cues and temporal dynamics are beneficial for RGB data processing, hence the higher performance with 3D-CDC-ST.

Additionally, we evaluated various values of theta to assess the impact of the 3D-CDC on performance. Notably, the results indicate that when compared to the 3D vanilla convolution ($\theta$ = 0), the 3D-CDC-ST ($\theta$ = 0.6) significantly enhances the performance in RGB data, while the 3D-CDC-T ($\theta$ = 0.6) improves the performance for depth data. These findings suggest that the optimal trade-off between the vanilla 3D and 3D-CDC is achieved at $\theta$ = 0.6.

\subsubsection{\textbf{Effect of Residual Connection}}

In order to thoroughly investigate the effectiveness of residual connections within our proposed MFST-Base and MFST-Large model across various stages, we conducted experiments that included or excluded this component on RGB data. As shown in Table 5, the model with residual connections outperforms the model without, providing evidence of the effectiveness of applying residual connections among different factorized spatio-temporal stages. These connections allow for the direct flow of information from one stage to the subsequent stage, enabling the network to reuse earlier spatio-temporal features smoothly.

\begin{table}[]
\centering
\caption{Recognition accuracy comparison of our proposed MFST-Base and MFST-Large model with and without residual connection.}
\begin{tabular}{lll}
\hline
Res. Connect     & MFST-Base  & MFST-Large  \\ \hline
$\times$           & 90.2              & 91.1               \\
$\checkmark$       & 90.5              & 92.1               \\ \hline
\end{tabular}
\end{table}

\subsubsection{\textbf{Effect of Weight-shared Multi-Scale Transformer}}

We conduct an ablation study to evaluate the effectiveness of the proposed weight-shared multi-scale Transformer implemented in the WMS-Trans block. Our results are presented in Table 6. We compared the performance of the vanilla Transformer (V-Trans) with that of the WMS-Trans, using the same layer number, head number, and dimension. Our results indicate that the WMS-Trans achieved superior performance compared to the V-Trans, with improvements of 1.1\% on RGB data and 1.6\% on depth data. This validates the efficiency of the WMS-Trans design, which enables it to capture more complex patterns and dependencies and learn richer feature representations. 

\begin{table}[]
\centering
\caption{Recognition accuracy comparison between vanilla Transformer and weight-shared Transformer.}
\begin{tabular}{lll}
\hline
Trans Type & RGB(\%) & DEP(\%) \\ \hline
V-Trans    & 91.0        & 92.2            \\
WMS-Trans  & 92.1        & 93.8            \\ \hline
\end{tabular}
\end{table}

\subsubsection{\textbf{Qualitative Analysis}}

\begin{figure}[htbp]
\centering
\includegraphics[height=4.5cm]{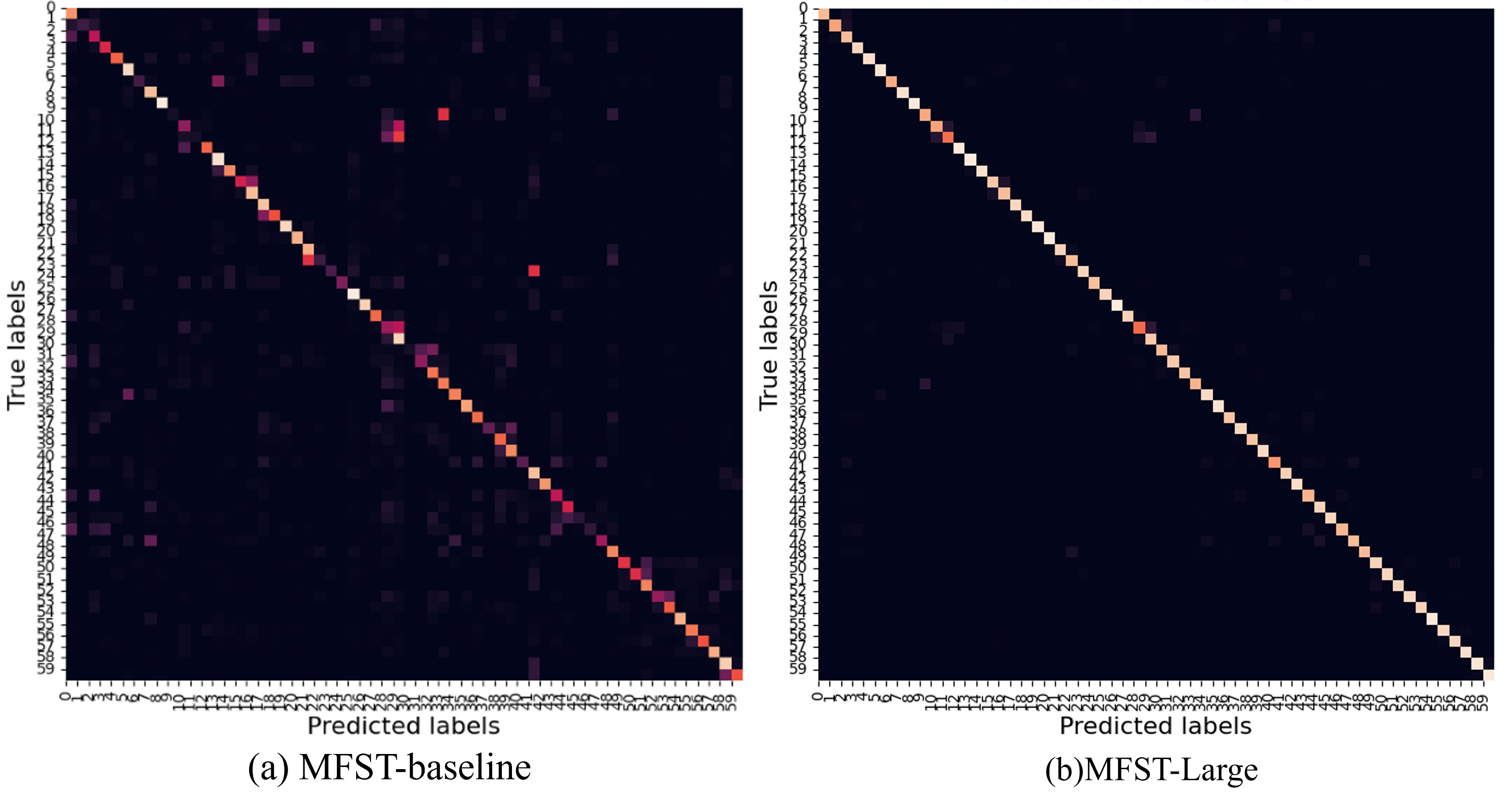}
\caption{Confusion matrices for RGB modality and CS protocol on the NTU RGB-D dataset of MFST-BM and MFST-Large models.}

\label{Fig6}
\end{figure}

The confusion matrices for the RGB modality and CS protocol on the NTU RGB-D dataset are shown in Figure \ref{Fig6} for the MFST baseline model and MFST-Large models. Figure \ref{Fig6}(a) reveals some misclassifications, such as "\#10 clapping" being confused with "\#34 rub two hands," "\#25 reach into pocket" being confused with "\#42 staggering," and "\#30 type on a keyboard" being confused with "\#12 writing." However, the confusion matrix of the MFST-Large model shows significant improvements in these subtle actions. This indicates the effectiveness of our proposed strategies for categorizing subtle actions through the learned hierarchical local fine-grained and global coarse-grained spatio-temporal features.

\vspace{-0.1cm}

\section{Conclusion}

We propose a multi-stage factorized spatio-temporal learning model for RGB-D action and gesture recognition that aims to individually capture spatial and temporal features in each stage while also modeling their dependencies without additional recoupling operations. Additionally, we introduce a lightweight CDC Stem that can efficiently capture coarse- and fine-grained temporal patterns. At each factorized spatio-temporal learning stage, our CNN-Transformer hybrid block and weight-shared Transformer block effectively capture multi-scale spatial and temporal features. As a result, our model achieves new state-of-the-art performance on both RGB-D action and gesture recognition benchmarks, including NTU RGB-D and IsoGD. These results demonstrate the effectiveness of our proposed multi-stage factorized spatio-temporal representation. In the future, we aim to leverage the multi-modality fusion method to represent more comprehensive spatio-temporal information from different modalities in an end-to-end manner.


\bibliographystyle{IEEEtran}

\bibliography{sample-base}
\end{document}